\documentclass[letterpaper, 10pt, conference]{style/ieeeconf}    

\makeatletter
\let\NAT@parse\undefined
\makeatother

\usepackage{dblfloatfix}
\usepackage[numbers, sectionbib,sort]{natbib}
\usepackage{bm}
\usepackage{gensymb}
\usepackage{xcolor}
\usepackage{graphicx}
\usepackage{amsmath}
\usepackage{amssymb}
\usepackage{subcaption}
\usepackage{amsfonts}
\usepackage{siunitx}
\usepackage{booktabs}
\usepackage{makecell}
\usepackage{multirow}
\usepackage{upgreek}
\usepackage[font=small]{caption}
\usepackage[export]{adjustbox}
\usepackage{tikz}
\usepackage{tabularx}
\usepackage{svg}
\svgsetup{inkscapepath=svgsubdir}
\usepackage{sidecap} \sidecaptionvpos{figure}{c}
\usepackage[hidelinks]{hyperref}

\usepackage[nameinlink, capitalize]{cleveref}
\usepackage{color}
\definecolor{CommentPink}{rgb}{1,0.2,0.5}
\definecolor{CommentBlue}{rgb}{0,0,1}
\definecolor{CommentGreen}{rgb}{0,1,0}

\Crefname{section}{Sec.}{Sec.}
\Crefname{equation}{Eq.}{Eq.}

\usepackage[printonlyused,withpage,nolist,nohyperlinks]{acronym}

\IEEEoverridecommandlockouts                       
\title{\LARGE \bf STAIR: Semantic-Targeted Active Implicit Reconstruction}

\author{Liren Jin \and Haofei Kuang \and Yue Pan \and Cyrill Stachniss \and Marija Popovi\'{c}%
\thanks{This work has been fully funded by the Deutsche Forschungsgemeinschaft (DFG, German Research Foundation) under Germany's Excellence Strategy, EXC-2070 -- 390732324 (PhenoRob). All authors are with the Institute of Geodesy and Geoinformation, University of Bonn. Cyrill Stachniss is also with Lamarr Institute for Machine Learning and Artificial Intelligence.
Corresponding: \texttt{ljin@uni-bonn.de}.}
}

\begin{document}
\maketitle
\thispagestyle{empty}
\pagestyle{empty}
\begin{abstract}
Many autonomous robotic applications require object-level understanding when deployed. Actively reconstructing objects of interest, i.e. objects with specific semantic meanings, is therefore relevant for a robot to perform downstream tasks in an initially unknown environment. In this work, we propose a novel framework for semantic-targeted active reconstruction using posed RGB-D measurements and 2D semantic labels as input. The key components of our framework are a semantic implicit neural representation and a compatible planning utility function based on semantic rendering and uncertainty estimation, enabling adaptive view planning to target objects of interest. Our planning approach achieves better reconstruction performance in terms of mesh and novel view rendering quality compared to implicit reconstruction baselines that do not consider semantics for view planning. Our framework further outperforms a state-of-the-art semantic-targeted active reconstruction pipeline based on explicit maps, justifying our choice of utilising implicit neural representations to tackle semantic-targeted active reconstruction problems.
\end{abstract} 
\section{Introduction} \label{S:introduction}
Active 3D reconstruction is relevant for many autonomous robot tasks in unknown environments~\citep{chen2011ijrr}. In various applications, including search and rescue, robot manipulation, and precision agriculture, the ability to extract accurate information about the geometry and appearance of objects of interest, i.e. objects with specific semantic meanings, is crucial for performing downstream tasks involving object-level understanding. A key challenge in such scenarios is planning a view sequence to get the most informative measurements targeting the objects of interest given a limited measurement budget, e.g. operation time or total number of measurements to be integrated.

\begin{figure}[!t]
\centering
  \begin{subfigure}[]{0.95\columnwidth}
  \includegraphics[width=\columnwidth]{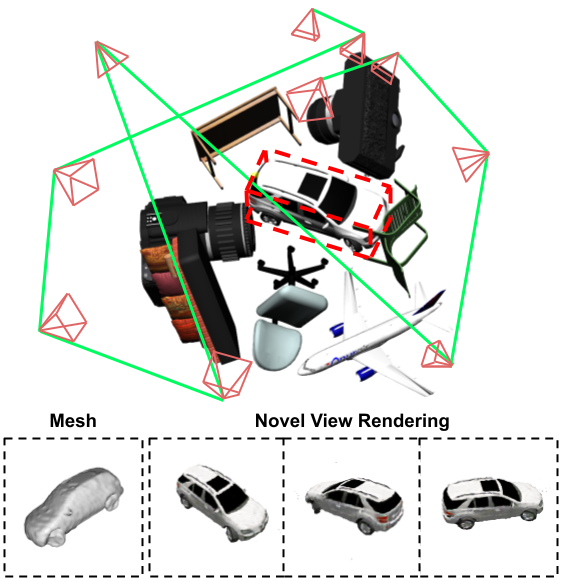}
  \end{subfigure}
    \caption{Our novel active implicit reconstruction approach targets an object of interest (car) in an unknown environment. We incorporate semantics and uncertainty estimation into our pipeline, enabling view planning to acquire information about the object in a targeted way. 
    The red bounding box identifies the target object. The green line shows the planned path, with pyramids indicating view frustums. By integrating semantics into our implicit neural representation, we extract mesh and render novel views only for the object of interest as exemplified in the bottom row.} \label{F: teaser}
\vspace{-0.5cm}
\end{figure}

In this work, we address the problem of actively reconstructing objects of one or multiple interesting semantic classes in an initially unknown 3D environment using posed RGB-D camera measurements.
Given a limited measurement budget, our goal is to obtain accurate 3D representations of the objects of interest by positioning a robotic camera online, i.e. during a mission, as shown in \cref{F: teaser}. Most existing approaches for active reconstruction~\citep{isler2016icra, palazzolo2018drones, naazare2022ral, pan2022eccv, yan2023ral, zhan2022activermap, jin2023iros, lee2022ral, he2023active, niko2023icra} aim at reconstructing the whole scene, without distinguishing between the observed objects. Since they do not incorporate semantics within planning pipelines, these methods cannot target specific objects of interest.

Recently, implicit neural representations~\citep{mescheder2019cvpr, park2019cvpr}, e.g. Neural Radiance Fields (NeRFs)~\citep{mildenhall2020eccv}, are attracting increasing attention as a compact form for dense scene representation. Follow-up works~\citep{sun2022cvpr, oechsle2021iccv, chen2022eccv, mueller2022tog} address the inherent training inefficiency of implicit neural representations by introducing hybrid structures, which learn scene attributes using sparse feature voxel grids combined with shallow multi-layer perceptrons (MLPs). This efficient structure enables deploying implicit neural representations in online robotic tasks~\citep{zhu2022cvpr, zhang2023corl, zhong2023icra}, while preserving their continuous representation capabilities. In this work, we exploit hybrid implicit neural representations as our map representation for semantic-targeted active implicit reconstruction. 

Active implicit reconstruction is an advancing research field. State-of-the-art works adopt next-best-view planning strategies to find the most informative measurements for training implicit neural representations. While showing promising results, these methods~\citep{pan2022eccv, yan2023ral, zhan2022activermap, jin2023iros, lee2022ral, he2023active, niko2023icra} only focus on uniformly reconstructing global scenes. They do not account for semantic information to distinguish objects of interest and reconstruct them in an adaptive, targeted way. In the context of semantics, recent works~\citep{zhi2021iccv, vora2022tmlr, bhalgat2023neurips, siddiqui2023cvpr} propose integrating 2D semantic labels into implicit neural representations to enhance semantic understanding capabilities. These approaches show accurate and consistent semantic rendering at novel views via multi-view learning. However, they have not been used for active reconstruction applications. To bridge the gap between active reconstruction and semantic implicit neural representations, we propose a new framework that enables guiding view planning toward objects of interest in an unknown environment.

Our main contribution is a novel framework, STAIR, for semantic-targeted active implicit reconstruction. Given posed RGB-D measurements and corresponding 2D semantic labels, our approach utilises implicit neural representations to learn occupancy, colour, and semantic fields associated with the scene. 
A key component of our approach is a new utility function for next-best-view planning using semantic implicit neural representations, which enables trading off between exploring the unknown environment and exploiting information about objects of interest as they are discovered. 

We make the following three claims: (i) our STAIR framework shows better performance in terms of reconstructed mesh and RGB rendering quality compared to pure exploration and heuristic baselines that do not consider semantics for view planning; (ii) our method outperforms a state-of-the-art semantic-targeted active reconstruction system using explicit map representations both in mapping and planning aspects; and
(iii) our utility function for planning balances between exploration and exploitation to handle challenging scenes containing many occlusions.
To support reproducibility and future research, our simulation environment and implementation will be released at: \url{https://github.com/dmar-bonn/stair}.

\section{Related Work} \label{S:related_work}
Our approach lies at the intersection of active reconstruction using semantics and implicit neural representations. In this section, we overview related work in these fields.

\subsection{Semantic-Targeted Active Explicit Reconstruction} \label{SS:semantic-targeted_exploration} 
Semantic understanding is crucial for many autonomous robotic tasks in unknown environments. Recent advancements in deep learning-based semantic segmentation facilitate the seamless integration of semantic understanding onboard robotic systems~\citep{juana2022}. In the context of active reconstruction, several works propose integrating semantics into explicit maps to enable semantic-targeted view planning.

\citet{papatheodorou2023icra} use an occupancy voxel map to model the background for exploring unknown environments. Once objects of predefined interesting semantic classes are found, they use adaptive-resolution octree-based signed distance function mapping to reconstruct the objects in detail.
\citet{lehnert2019iros} design a 3D camera array to obtain multiple measurements from different perspectives. The objects of interest detected in each measurement are used to calculate the gradient indicating the most likely direction of movement to observe them.
\citet{burusa2023efficient} calculate the expected information gain based on the confidence score of a voxel belonging to interesting semantic classes.
Similar to our problem setup, \citet{tobias2021iros} propose a semantic-targeted active explicit reconstruction system based on occupancy voxel maps and apply it to reconstruct fruits in agricultural robotics applications. To guide targeted next-best-view planning, they assign higher utility for candidate views that observe more unknown voxels close to already detected objects of interest. 

Our approach shares the same idea of using semantic information to conduct view planning towards objects of interest. However, different from previous works that rely on discrete explicit maps, we exploit recent advances in implicit neural representations to improve the reconstruction quality.

\subsection{Active Implicit Reconstruction} \label{SS:active_implicit_reconstruction} 
Implicit neural representations are a powerful tool for 3D reconstruction due to their continuous representation capabilities. Recent work has explored how to exploit these benefits in active reconstruction settings.

\citet{pan2022eccv} model the radiance field as Gaussian distribution and actively collect images by evaluating the reduction of uncertainty assuming new inputs at candidate views. Exploiting fast rendering of Instant-NGP~\citep{mueller2022tog}, \citet{niko2023icra} train an ensemble of NeRF models for a single scene and measure uncertainty as the variance of the ensemble's prediction, which is used to conduct next-best-view selection. \citet{jin2023iros} incorporate uncertainty estimation into image-based neural rendering to predict rendering uncertainty at novel views, enabling mapless next-best-view planning. Leveraging the differentiability of the implicit neural representations, \citet{yan2023ral} optimise next-best-view generation towards views with high uncertainty. Following a different paradigm, \citet{pan2024icra} utilise a view number prediction network to predict the number of views required to reconstruct a specific unknown object using NeRF, allowing for one-shot view sequence generation without online replanning.

Our work follows these lines by using implicit neural representations for active reconstruction. Different from previous methods that uniformly reconstruct a scene or an object, our approach integrates semantic understanding into an implicit neural representation to achieve semantic-targeted active implicit reconstruction.

\begin{figure*}[!t]
\centering
  \includegraphics[width=0.95\textwidth]{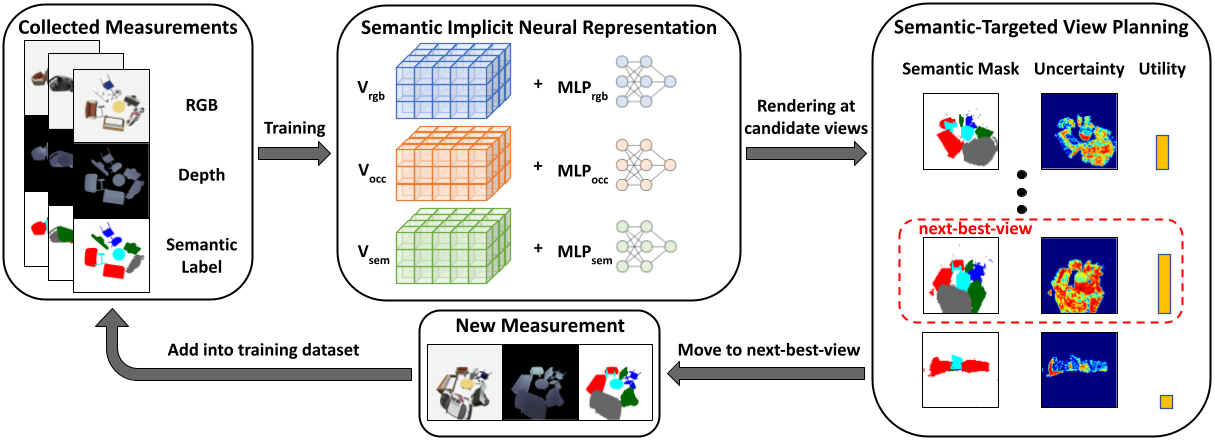}
  \caption{Overview of our proposed framework, STAIR. We incrementally train our semantic implicit neural representation using posed RGB-D measurements and their 2D semantic labels. After training, we render semantics and uncertainty at sampled candidate views. For planning, our utility function considers both overall view uncertainty and the uncertainty from objects of interest. We select the candidate view with the highest utility value as our next measurement location. We iterate between map representation training and view planning until a maximum allowable number of measurements is reached.}
  \label{F: framework}
\vspace{-0.5cm}
\end{figure*}

\subsection{Semantics in Implicit Neural Representations} \label{SS:semantic_implicit_neural_representations}

Recent works propose lifting 2D semantic information into 3D to generate a consistent semantic field by exploiting the multi-view consistency from learning implicit neural representations. \citet{zhi2021iccv} extend vanilla NeRF to jointly encode the semantics along with the scene appearance and geometry. Their results show multi-view consistent and smooth semantic rendering at novel views, even given sparse or noisy 2D semantic labels as supervision signals. \citet{siddiqui2023cvpr} and \citet{bhalgat2023neurips} further incorporate instance segmentation into implicit neural representations. 
Vora et al.~\cite{vora2022tmlr} train a 3D network to convert a learned density field into a semantic field, which generalises across scenes.

In contrast to previous approaches for generating semantic implicit neural representations, \citet{kelly2023icra} use semantic information to train NeRFs in a targeted way. To reconstruct objects of interest in the scene at higher quality, they propose a denser sampling of training examples around these objects based on semantic segmentation. DietNeRF~\citep{jain2021iccv} proposes a semantic consistency loss to regularise rendering from arbitrary views, encouraging consistent high-level semantics. This additional loss alleviates the degenerate performance commonly observed in NeRF training with sparse views.

While semantics offer rich scene understanding capabilities in implicit neural representations, they have not yet been applied for active implicit reconstruction problems. We bridge this gap by introducing a framework for semantic-targeted active reconstruction based on implicit neural representations. Our approach is applicable for similar problems tackled by current methods using active explicit reconstruction to target objects of interest in unknown environments~\citep{tobias2021iros, burusa2023efficient,papatheodorou2023icra}. However, we exploit the advantages of underlying implicit neural representations to further improve the reconstruction quality.
 
\section{Our Approach} \label{S:our_approach}

We propose STAIR, a novel framework for semantic-targeted active implicit reconstruction in autonomous robotics tasks. An overview of our framework is shown in~\cref{F: framework}. Our goal is to actively reconstruct objects of interest in an initially unknown environment using a robot equipped with a RGB-D camera. We utilise an implicit neural representation consisting of sparse feature voxel grids and MLPs as our map representation. Given collected posed RGB-D measurements and corresponding semantic labels, we incrementally train our map representation to model the occupancy probability, colour, and semantic information in continuous 3D space. To guide semantic-targeted view planning, we sample candidate views in a predefined action space and evaluate the utility of each view based on uncertainty estimates from the occupancy distribution and semantic rendering. The candidate view with the highest utility value is selected as the location for the next measurement. We iterate between training and planning until a maximum allowable number of measurements is reached. 

\subsection{Semantic Implicit Neural Representation}\label{SS:semantic_implicit_neural_representation}
Similar to DVGO~\citep{sun2022cvpr}, our map representation consists of sparse feature voxel grids and MLPs to balance representation capabilities and training efficiency.
We maintain features for different modalities of the scene: spatial occupancy (occ), RGB colour (rgb), and semantics (sem), in three voxel grids $\mathbf{V}_{\mathrm{occ}}$, $\mathbf{V}_{\mathrm{rgb}}$, and $\mathbf{V}_{\mathrm{sem}}$,
respectively. For any point in space, we can query its modality feature by trilinear interpolation in the corresponding voxel grid expressed as: 
\begin{align}
\mathbf{f}_{\mathit{m}} = \mathrm{interp}(\mathbf{x}, \mathbf{V}_{\mathit{m}}): (\mathbb{R}^{3} \times \mathbb{R}^{T_{\mathit{m}} \times H \times W \times L}) \rightarrow \mathbb{R}^{T_{\mathit{m}}} \, , 
\label{E:interpolation}
\end{align}
where $\mathit{m} \in \{\mathrm{occ}, \mathrm{rgb}, \mathrm{sem} \}$, $\mathbf{f}_{\mathit{m}} \in \mathbb{R}^{T_{\mathit{m}}}$ is the queried modality feature vector at position $\mathbf{x} \in \mathbb{R}^3$, $\mathbf{V}_{\mathit{m}}$ is the feature voxel grid of corresponding modality with $T_{\mathit{m}}$ feature channels, and $H$, $W$, $L$ are the spatial resolution dimensions.

The queried modality features at point $\mathbf{x}$ are then interpreted by modality-specific MLPs into per-point occupancy probability $o(\mathbf{x}) = \mathrm{MLP}_{\mathrm{occ}}(\gamma(\mathbf{x}), \, \mathbf{f}_{\mathrm{occ}}) \in \left [ 0,\,1 \right ]$, RGB colour $\mathbf{c}(\mathbf{x}) = \mathrm{MLP}_{\mathrm{rgb}}(\gamma(\mathbf{x}), \, \mathbf{f}_{\mathrm{rgb}}) \in \left [ 0,\,1 \right ]^3 $, and semantic probability vector $\mathbf{s}(\mathbf{x}) = \mathrm{MLP}_{\mathrm{sem}}(\gamma(\mathbf{x}), \, \mathbf{f}_{\mathrm{sem}}) \in \left [ 0,\,1 \right ]^{P}$, with $P$ as the number of total semantic classes. We use a positional encoding function~\citep{mildenhall2020eccv} $\gamma : \mathbb{R}^3 \rightarrow \mathbb{R}^{21}$ to map position $\mathbf{x}$ into a higher-dimensional space. Note that we do not consider view-dependent colour emission in this work.

\subsection{Training of Map Representation}\label{SS:training_of_map_representations}
Our map representation is updated online during a mission. Given a set of posed RGB-D measurements obtained by the robot camera and their semantic labels, we jointly train our feature voxel grids and MLPs using differentiable volume rendering~\citep{mildenhall2020eccv}.
To render colour, depth, and semantics for a ray $\mathbf{r}$ cast from a measurement view, we uniformly sample $N$ points $\mathbf{x}_{i \in \left\{1, 2, \ldots, N \right\}}$ along the ray with $d(\mathbf{x}_i)$ as the depth value from the sampling point $\mathbf{x}_{i}$ to its view origin. Following UNISURF~\citep{oechsle2021iccv}, occupancy-based volume rendering for predicted colour $\hat{C}(\mathbf{r})$, depth $\hat{D}(\mathbf{r})$, and semantic probability $\hat{S}(\mathbf{r})$ observed from ray $\mathbf{r}$ is given by:
\begin{align}
\hat{C}(\mathbf{r}) = \sum_{i=1}^{N} w(\mathbf{x}_i) \, \mathbf{c}(\mathbf{x}_i) \, , \\
\hat{D}(\mathbf{r}) = \sum_{i=1}^{N} w(\mathbf{x}_i) \, d(\mathbf{x}_i) \, , \\
\hat{S}(\mathbf{r}) = \sum_{i=1}^{N} w(\mathbf{x}_i) \, \mathbf{s}(\mathbf{x}_i)\, , \label{E: semantic_rendering}
\end{align}
with:
\begin{equation}
w(\mathbf{x}_i) = o(\mathbf{x}_i)\,T(\mathbf{x}_i) \, , \quad T(\mathbf{x}_i) = \prod_{j<i} \left(1-o(\mathbf{x}_j)\right)\, ,
\label{E: weight}
\end{equation}
where $w(\mathbf{x}_i)$ is the weight of modality value at $\mathbf{x}_i$ and $T(\mathbf{x}_i)$ is accumulated transmittance, indicating the probability of ray reaching $\mathbf{x}_i$ without being blocked by built surfaces. 

We supervise the training using the loss terms: 
\begin{align}
  \mathcal{L}_{\mathrm{rgb}} &= \sum_{\mathbf{r} \in \mathcal{R}} \left \| C(\mathbf{r}) - \hat{C}(\mathbf{r}) \right \|_{2} \, ,
  \label{E:rgb_loss_function} \\
  \mathcal{L}_{\mathrm{depth}} &= \sum_{\mathbf{r} \in \mathcal{R}} \left \| D(\mathbf{r}) - \hat{D}(\mathbf{r}) \right \|_{1} \, ,
  \label{E:depth_loss_function} \\
  \mathcal{L}_{\mathrm{sem}} &= \sum_{\mathbf{r} \in \mathcal{R}} \mathrm{CE} ((S(\mathbf{r}), \, \hat{S}(\mathbf{r})) \, ,
  \label{E:semantic_loss_function}
\end{align}
where $C(\mathbf{r})$, $D(\mathbf{r})$, and $S(\mathbf{r})$ are the recorded colour, depth, and semantic label respectively of ray $\mathbf{r}$ in the measurements, CE refers to the cross entropy loss and $\mathcal{R}$ denotes the set of rays in the training batch.
The total training loss is then: 
\begin{equation}
    \small{
     \mathcal{L} = \lambda_1 \mathcal{L}_{\mathrm{rgb}} +\lambda_2 \mathcal{L}_{\mathrm{depth}} + \lambda_3 \mathcal{L}_{\mathrm{sem}} \, ,
    }
     \label{E:loss_function}
\end{equation}
with the factors $\lambda_1, \lambda_2, \lambda_3$ balancing the weight of each term in the loss function.

We incrementally train our map representation for a constant number of iterations when a new measurement arrives. To avoid overfitting to the latest measurement, we collect our training batch $\mathcal{R}$ for each training iteration from both previous measurements and the latest measurement. We assign the probability of sampling each training ray example as being inversely proportional to its total sampled time to ensure uniform sampling across the whole training dataset. After training, our map representation is used for semantic-targeted view planning, introduced next.

\subsection{Semantic-Targeted View Planning}\label{SS:semantic-targeted_view_planning}
A key aspect in our framework is a utility function that adaptively guides view planning by trading off between exploration and exploitation. We first introduce our sampling strategy for generating candidate views and then elaborate on how we calculate utility values for next-best-view selection. 

To generate candidate views, we adopt a two-stage sampling strategy. We first uniformly sample $N_\mathrm{uni}$ candidate views on the hemispherical action space. We evaluate the individual utility at each view and select the views of top $K$ utility values. We then resample $N_\mathrm{re}$ new candidate views around each of these views to obtain a fine-grained utility evaluation. Finally, the candidate view with the highest utility value is selected as the next-best-view.

Our utility quantification requires uncertainty estimates and semantic rendering. Uncertainty estimation indicates parts of the scene that are unexplored or still not well-reconstructed. On the other hand, semantic rendering provides masks to distinguish objects of interest, allowing for view selection in a targeted way. We derive the uncertainty estimates from our trained occupancy field. For a candidate view $v_k$, we sample $N_\mathrm{pt}$ points on each of $N_\mathrm{ray}$ rays cast from the view. We define the uncertainty at each sampling point $\mathbf{x}_i$ as its entropy:
\begin{equation}
\small{
   H_\mathrm{pt}(\mathbf{x}_i) = -o(\mathbf{x}_i) \, \mathrm{ln}(o(\mathbf{x}_i)) - \bar{o}(\mathbf{x}_i) \, \mathrm{ln}(\bar{o}(\mathbf{x}_i)) \, ,
   }
\end{equation}
where $\bar{o} = 1-o$ is the complementary occupancy probability. Note that we do not consider the entropy of sampling points behind the built object surface. Thus, the total entropy along a ray $\mathbf{r}$ is:
\begin{equation}
\small{
   H_\mathrm{ray}(\mathbf{r}) = \sum_{i=1}^{N_\mathrm{pt}} T(\mathbf{x}_i) \, H_\mathrm{pt}(\mathbf{x}_i) \,, 
   }
\end{equation}
 where $T$ is the accumulated transmittance term introduced in \cref{E: weight}. 
The total uncertainty rendered at view $v_k$ is:
\begin{equation}
\small{
    U_\mathrm{er}(v_k) = \sum_{i=1}^{N_\mathrm{ray}} H_\mathrm{ray}(\mathbf{r}_i) \, ,
    }
    \label{E: exploration}
\end{equation}
which we define as our exploration (er) score. This term does not distinguish between the uncertainty values associated with different objects. Instead, it quantifies the total uncertainty at a view.
To account for objects of interest based on their semantic meaning, we apply a mask to the uncertainty according to whether or not the objects are relevant for semantic-targeted active planning:
\begin{align}
    U_\mathrm{et}(v_k) = \sum_{i=1}^{N_\mathrm{ray}} H_\mathrm{ray}(\mathbf{r}_i) \label{E: exploitation}\delta(\mathbf{r}_i) \, ,  \\
    \delta(\mathbf{r}_i) = 
    \begin{cases}
        1 & \text{if } \mathrm{argmax}(\hat{S}(\mathbf{r}_i)) \in \mathcal{T}\\
        0 &  \text{otherwise}
    \end{cases}
    \, ,
\end{align} 
where $\hat{S}(\mathbf{r}_i)$ is the predicted semantic probability vector obtained using \cref{E: semantic_rendering} and $\mathcal{T} \subseteq \{1, 2, \ldots, P\}$ is a set of identifiers for the interesting semantic classes.
We denote the sum of pixelwise uncertainty from the objects of interest as our exploitation (et) score, which guides view planning towards target objects. 

To trade off between exploring the unknown
environment and exploiting information about objects of
interest as they are discovered, we compute the utility value of a candidate view as the sum of exploitation and weighted exploration score, with $\varepsilon$ as the weight factor:
\begin{equation}
\small{
    U(v_k) = U_\mathrm{et}(v_k) + \varepsilon U_\mathrm{er}(v_k) \, .
    }
    \label{E: utility}
\end{equation}

\section{Experimental Results} \label{S:experimental_evaluation}

\begin{figure}[!h]
\centering
  \begin{subfigure}[]{0.95\columnwidth}
\includegraphics[width=\columnwidth]{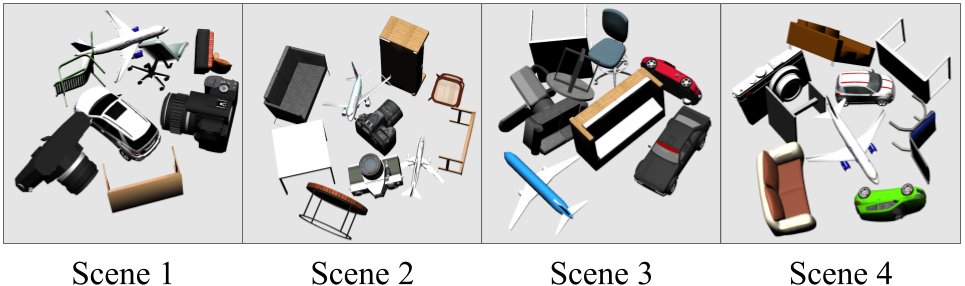}
  \end{subfigure}
  \caption{Four different scenes used in our main planning experiments. Our interesting semantic classes are: car for Scene 1, camera for Scene\,2, sofa for Scene 3, car and airplane for Scene 4.}
\vspace{-0.4cm}
  \label{F: simulation_scenes}
\end{figure}

\subsection{Experimental Setup}\label{SS:experimental setup}
\textbf{Simulator}. We spawn ShapeNet~\citep{chang2015shapenet} models of different semantic classes with random poses in Gazebo~\citep{gazebo} to build simulation scenes. We consider $7$ semantic classes in our simulator: car, airplane, sofa, chair, table, camera, and background. Four scenes used in the planning experiments are shown in~\cref{F: simulation_scenes}. All scenes consider a bounding box size of $3$\,m $\times$ $3$\,m $\times$ $3$\,m. We set our camera action space as a scene-centric hemisphere with $2$ m radius and camera views targeting the scene origin. All RGB-D measurements are at $400$\,px $\times$ $400$\,px resolution. 
To acquire the semantic labels, pre-trained semantic segmentation models can be applied; however, in this work, we use ground truth semantics from the simulator to focus on evaluating planning performance.

\textbf{Training Setup}. We use a grid size of $128 \times 128 \times 128$ for all three feature voxel grids. We set the feature channels as $T_{\mathrm{occ}} = 3$, $T_{\mathrm{rgb}} = 6$, and $T_{\mathrm{sem}} = 7$. The $\mathrm{MLP}_{\mathrm{rgb}}$ comprises two hidden layers with 128 channels, while $\mathrm{MLP}_{\mathrm{occ}}$ consists of two hidden layers with 32 channels. We simply use an identity mapping as $\mathrm{MLP}_{\mathrm{sem}}$ and no positional encoding for modelling semantics since the semantic field is smooth and exists in a low-frequency domain.
We set $\lambda_1 = 1.0, \lambda_2 = 0.1$, and $\lambda_3 = 1.0$ in \cref{E:loss_function}. For each training iteration, we use a batch size of $8000$ with $4000$ training examples from all previous measurements and $4000$ training examples from the current measurement. We train our map representation for $200$ steps before conducting view planning, which takes approximately $4$\,s with our PyTorch implementation running on a single NVIDIA RTX A5000 GPU.

\textbf{Planning Setup}.
For candidate view sampling introduced in \cref{SS:semantic-targeted_view_planning}, we set $N_\mathrm{uni} = 100$, $K=10$, and $N_\mathrm{re}=10$, giving a total of $200$ views. To render semantic and uncertainty maps at a candidate view, we use $N_\mathrm{ray} =80 \times 80$ and $N_\mathrm{pt} = 200$. One planning step takes around $2$\,s under this sampling and rendering configuration. The exploration weight $\varepsilon$ in \cref{E: utility} is $0.2$. We select car in Scene 1, camera in Scene 2, sofa in Scene 3, car and airplane in Scene 4 as the interesting classes for semantic-targeted active reconstruction. The maximum number of planning steps is set to $10$ for all experiments.

\textbf{Evaluation Metrics}.
We evaluate the reconstruction results with test view rendering performance and mesh quality. We report the peak signal-to-noise ratio (PSNR)~\citep{mildenhall2020eccv} as the rendering metric and use F1-score to measure overall mesh quality.
Since our goal is to reconstruct objects of interest, we only consider these objects in the metrics calculations. Hence, when rendering at test views or extracting meshes from our trained map representation, we only keep objects of interest by setting the occupancy probability of points with uninteresting semantic predictions to zero.

For calculating PSNR, we render colour images at $100$ uniformly distributed test views and compare the predictions with ground truth images. We average the PSNR over all test views as the final rendering metric. For mesh quality evaluation, we first extract the mesh of objects of interest from our trained occupancy field using Multiresolution IsoSurface Extraction~\citep{mescheder2019cvpr} with a threshold of $0.5$. We uniformly sample $10^6$ points on both the extracted mesh and the ground truth mesh. The precision is calculated as the fraction of points on the extracted mesh that are closer than a threshold distance to points on the ground truth mesh. Similarly, the completeness is the fraction of points on the ground truth mesh that match points on the extracted mesh within a threshold distance. We use $1$ cm as the threshold value for precision and completeness calculations. Finally, the F1-score is the harmonic mean of precision and completeness.

\subsection{Active Implicit Reconstruction} 
\label{SS:active_implicit_reconstruction}
\begin{figure*}[!h]
    \centering
    \includegraphics[width=0.95\textwidth]{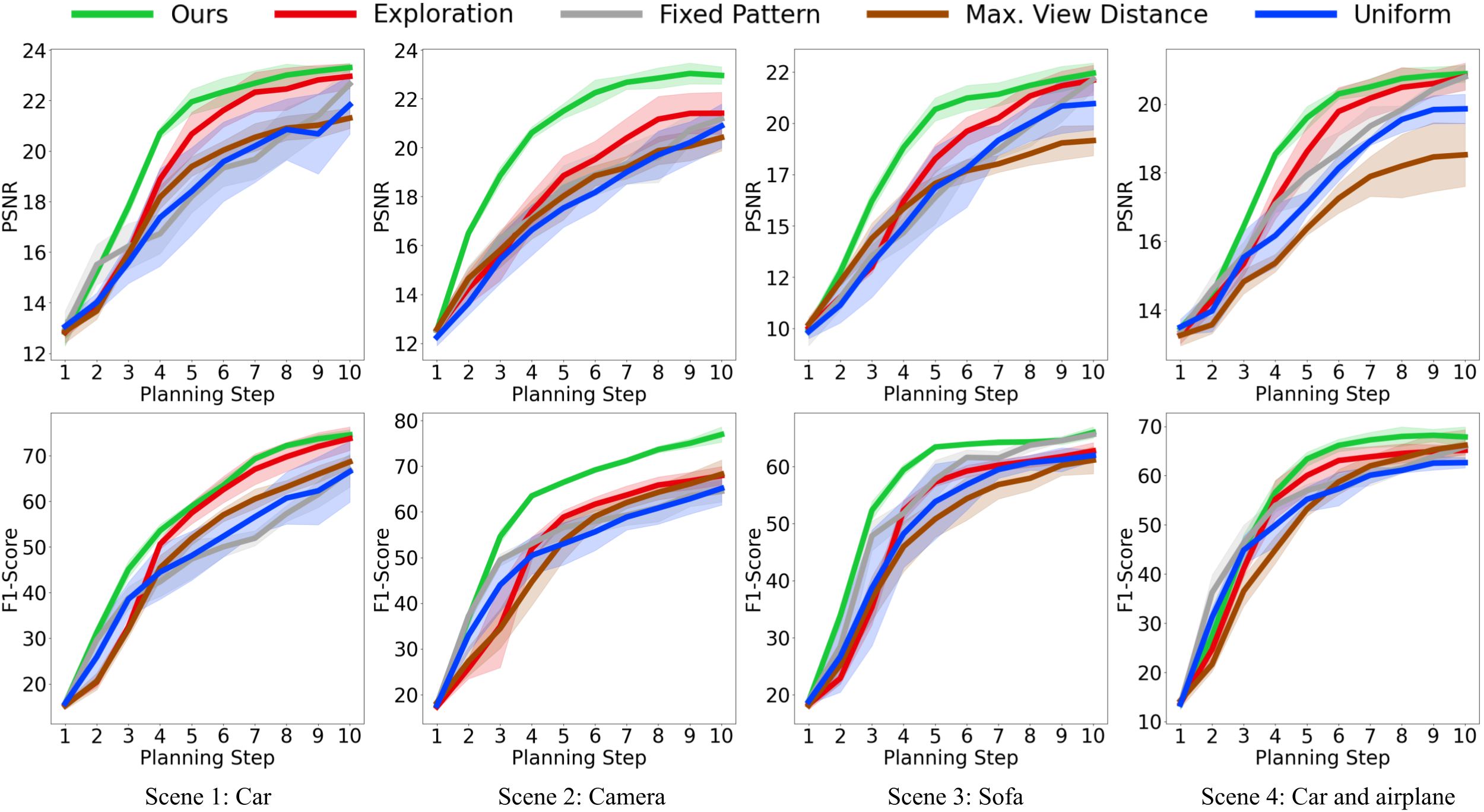}
    \caption{Comparison of reconstruction quality of objects of interest using different planning strategies in the four test scenes shown in \cref{F: simulation_scenes}. We report the average PNSR and F1-score at each planning step. Solid lines show means over $5$ trials and shaded regions indicate standard deviations. Our semantic-targeted approach exploits semantics in our implicit neural representation to achieve targeted view planning, leading to better and more stable reconstruction performance.} 
    \label{F: experiment1_quantitive}
\end{figure*}

\begin{figure*}[!h]
\centering
\includegraphics[width=0.95\textwidth]{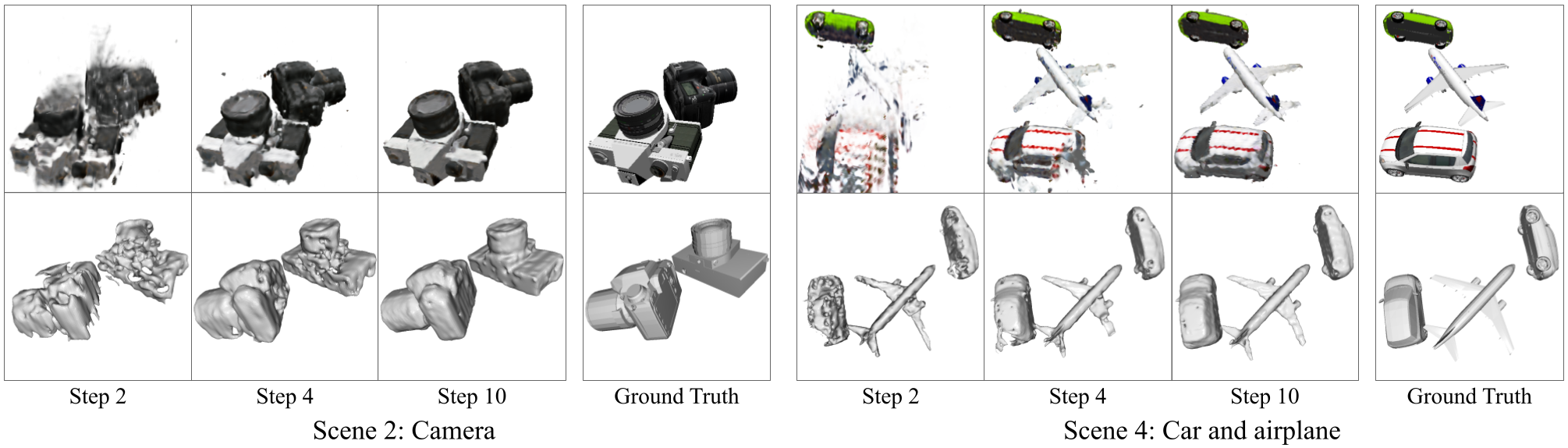}
\caption{Qualitative results using our framework showing how novel view rendering (top) and meshes (bottom) improve along planning steps during a mission. Our approach collects information about objects of interest in a targeted way to achieve high-quality reconstruction.}
\vspace{-0.5cm}
\label{F: experiment1_qualitative}
\end{figure*}

Our first experiment shows that our semantic-targeted view planning method achieves better reconstruction quality in terms of rendering performance and mesh quality compared to pure exploration and heuristic baselines that do not consider semantics. The map representations and training configurations are the same for all methods, hence the reconstruction quality differs purely as the consequence of collected measurements using different planning strategies. We consider the following planning methods:
\begin{itemize}
\item \textit{Ours}: selects the view with the highest utility value defined in~\cref{E: utility};
\item \textit{Exploration}: selects the view with the highest exploration score as calculated by~\cref{E: exploration};
\item \textit{Fixed Pattern}: follows the spiral pattern view sequence to cover the hemispherical action space;
\item \textit{Max. View Distance}: selects the view that maximises the view distance to all previously visited views;
\item \textit{Uniform}: selects a random view from uniformly sampled candidate views.
\end{itemize}

For all experiment runs, we start with a measurement from the top view and use different planning methods to select the next view to acquire a new measurement, which, together with all previous measurements, is used to train our map representation. We evaluate reconstruction performance after every planning step. For each test scene and planning method, we run $5$ trials and report the average PSNR and F1-score with standard deviations along the planning steps. 

The experiment results are given in~\cref{F: experiment1_quantitive}. Next-best-view planning guided by our approach shows steeper-rising metric curves, indicating more efficient reconstruction compared to baselines that do not consider semantic information. This verifies that our STAIR framework benefits from integrating semantics in an implicit neural representation to achieve semantic-targeted active reconstruction. Our approach has the lowest standard deviations across all scenes, indicating its robust performance. In~\cref{F: experiment1_qualitative}, we show two examples of how novel view rendering and object meshes improve along planning steps using our approach. 

\subsection{Comparison Against Active Explicit Reconstruction} \label{SS:comparison_againt_explicit_framework}

\begin{figure*}[!h]
    \centering  \includegraphics[width=0.95\textwidth]{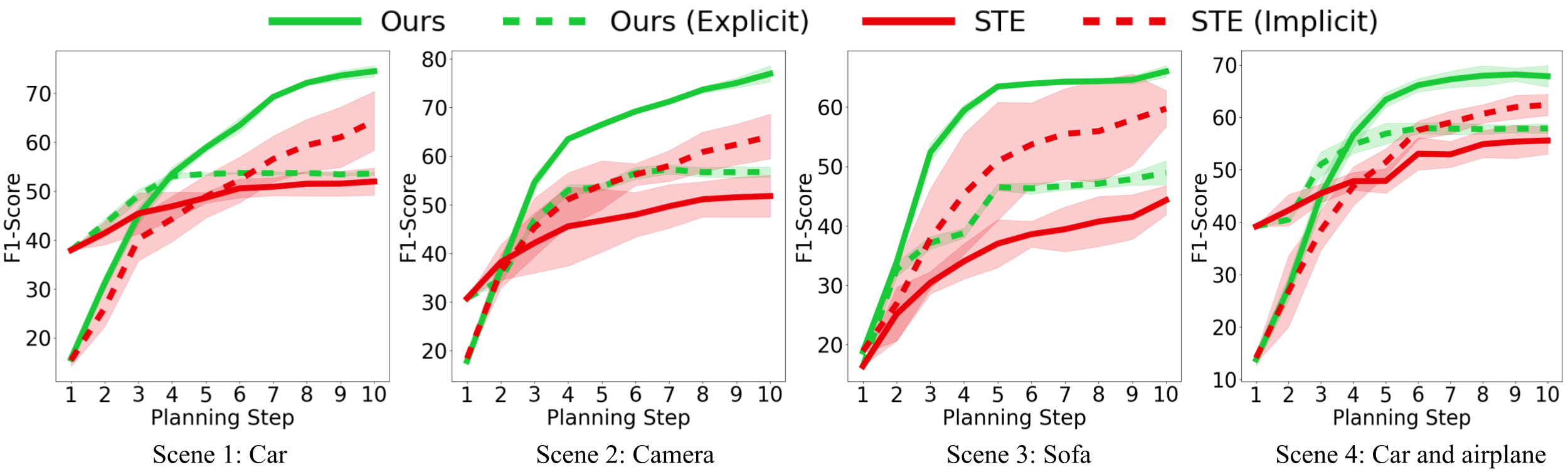}
\caption{Comparison of our STAIR framework against semantic-targeted active explicit reconstruction system $\textit{STE}$~\citep{tobias2021iros}. Dashed lines denote variants cross-validating the measurements collected by one active reconstruction system with the mapping method of the other. Same colour indicates mapping using the same measurements. The results confirm that our STAIR framework achieves superior performance compared to the explicit baseline. The performance gain originates due to the implicit neural representation used in our framework and our utility function for finding more informative measurements.}
    \vspace{-0.5cm}
\label{F: experiment2_quantitative}
\end{figure*}
\begin{figure}[!h]
\centering
\includegraphics[width=0.95\columnwidth]{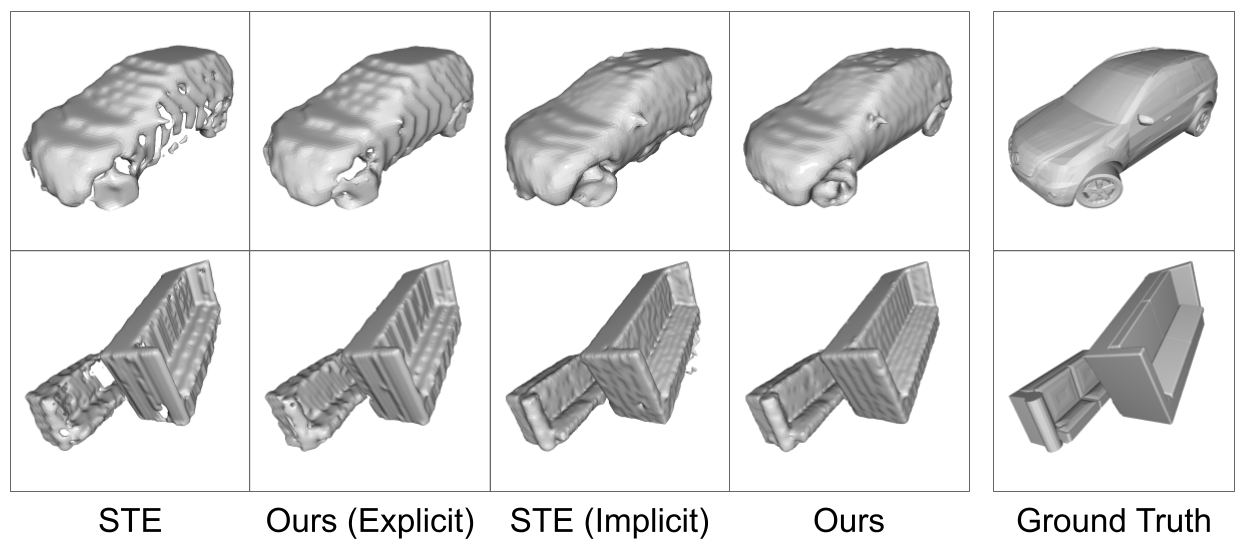}
\caption{Comparison of final mesh reconstructions. The meshes extracted from explicit map representations are limited by the discrete representation, containing holes and non-smooth surfaces. The implicit neural representation used in our framework results in better mesh quality, due to its continuous representation capabilities.}
\vspace{-0.5cm}
\label{F: experiment2_qualitative}
\end{figure}
In this experiment, we compare our STAIR framework against semantic-targeted active explicit reconstruction to show the advantages of using an implicit neural representation for our task. Specifically, we compare against the approach of~\citet{tobias2021iros}, which we denote as \textit{STE} to indicate semantic-targeted planning based on explicit map representations. 
\textit{STE} fuses RGB-D measurements and 2D semantic labels into an explicit semantic occupancy grid map and biases planning towards the objects of interest as they are built in the map by assigning higher utility to unknown voxels close to objects of interest.
For comparability, we use the same grid size of $128 \times 128 \times 128$ for their map.

To further investigate the sources of performance difference between our approach and \textit{STE}, we cross-validate these two active reconstruction frameworks by combining measurements collected by each framework with the other mapping system. After the online planning experiments, we fuse the measurements collected by our framework into an explicit occupancy map used in the \textit{STE} approach. We denote this combination as \textit{Ours (Explicit)}. The result of this combination indicates whether the performance gain originates from our view planning results. Similarly, we use the measurements collected by the \textit{STE} approach to train our implicit neural representation, which we denote as \textit{STE (Implicit)}. This combination exposes how different map representations influence the reconstruction performance. 

The results are shown in~\cref{F: experiment2_quantitative}. Our framework performs better than the $\textit{STE}$ method. The performance gain can be decomposed into two aspects. First, comparing $\textit{STE (Implicit)}$ and $\textit{STE}$ suggests that, given the same measurements, our implicit neural representation improves reconstruction quality compared to explicit occupancy mapping. This justifies the choice of using implicit neural representations in our active reconstruction framework. Second, as seen by comparing $\textit{Ours (Explicit)}$ and $\textit{STE}$, even when using explicit occupancy mapping, measurements acquired using our planning approach lead to better reconstruction quality. This indicates that our semantic-targeted view planning based on dense semantic and uncertainty rendering enables finding more informative views to reconstruct objects of interest. \cref{F: experiment2_qualitative} visualises the final extracted meshes using the four methods. Meshes extracted from our implicit neural representation show complete surfaces with more high-frequency details compared to those from explicit maps.

\subsection{Ablation Study} \label{SS:ablation_study}
\begin{figure}[!t]
\centering
\begin{subfigure}{0.95\columnwidth}
  \includegraphics[width=\columnwidth]{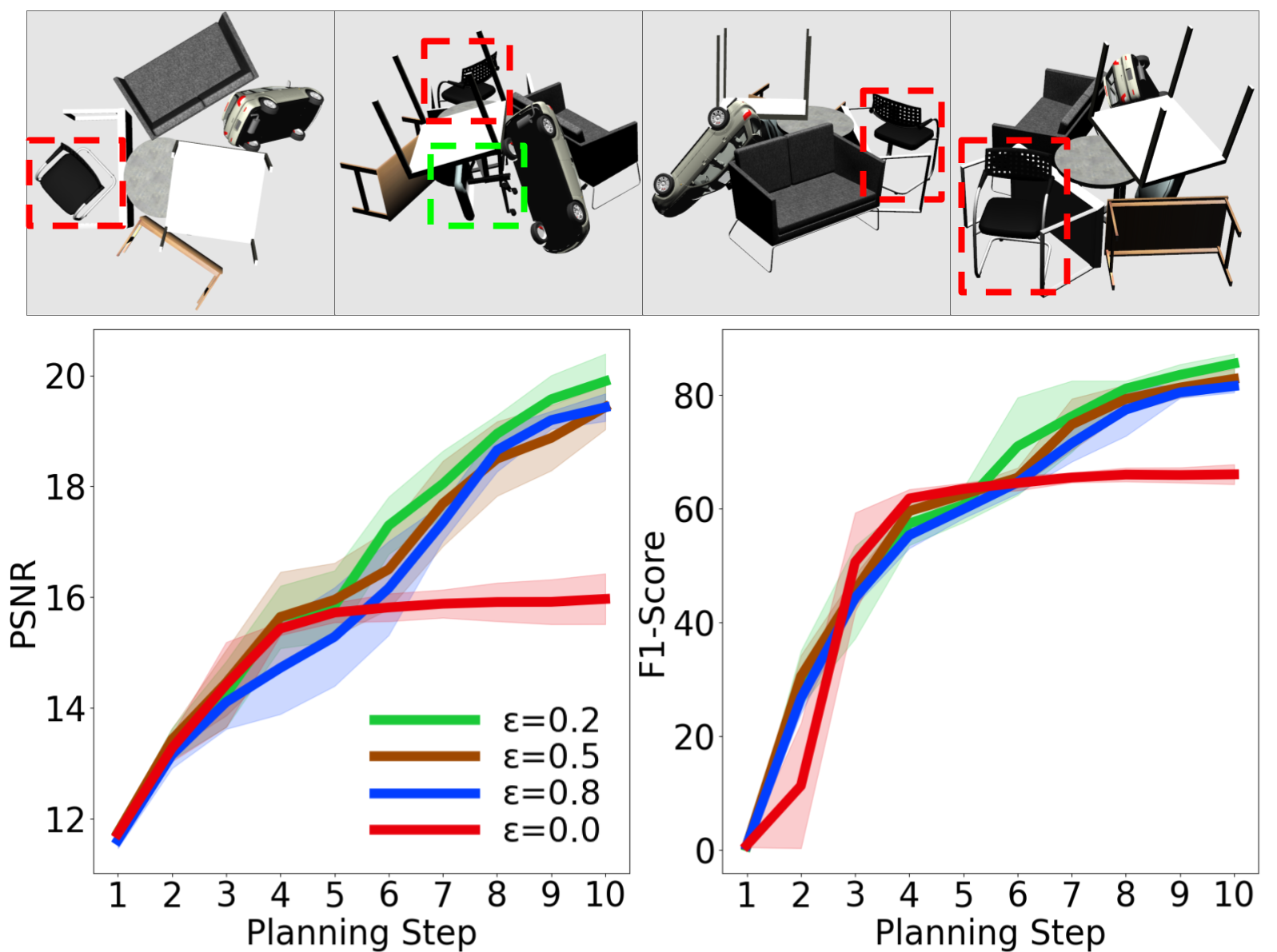}
\end{subfigure}
\caption{Top row: Test scene seen from different perspectives. One object of interest (red bounding box) can be easily detected; however, the second object of interest (green bounding box) is severely occluded by other objects and can only be observed from particular views. Bottom row: Semantic-targeted view planning using an exploitation term alone ($\varepsilon=0.0$) cannot explore to find both objects of interest. In contrast, our utility function balances between exploitation and exploration, leading to better active reconstruction performance in this challenging situation.}
\vspace{-0.6cm}
\label{F: ablation}
\end{figure}
The final experiment justifies our design choice for the utility function introduced in \cref{SS:semantic-targeted_view_planning}. 
We show that an exploration term is necessary for semantic-targeted view planning in an unknown environment. 
For this purpose, we design a challenging scene, as shown in~\cref{F: ablation}, where two objects of interest (chairs) are separated by other objects. We start from the top view, from which only one chair is seen and the other one is occluded. We compare the planning approach using the exploitation-only score in \cref{E: exploitation}, i.e. $\varepsilon = 0.0$, and our proposed utility function in \cref{E: utility} with $\varepsilon$ values of $0.2$, $0.5$, and $0.8$ to investigate the influence of varying the exploration term proportion.

\cref{F: ablation} compares the reconstruction performance in the test scene. Semantic-targeted view planning without exploration focuses only on already detected objects of interest. As a result, this planning strategy does not explore the unknown environment to find other potential objects of interest in the scene, leading to inferior overall reconstruction performance. In contrast, 
our approach trades off
between exploring the unknown environment and exploiting
information about objects of interest as they are discovered. The results indicate that a small exploration term is sufficient to achieve such behaviour, while up-weighting exploration deteriorates semantic-targeted view planning performance.

\section{Conclusions} \label{S:conclusions_and_future_work}
We presented STAIR, a novel framework for semantic-targeted active implicit reconstruction. Our approach exploits implicit neural representation with semantic understanding capabilities.
By combining uncertainty estimation and semantic rendering, our semantic-targeted view planning strategy gathers information about objects of interest in unknown environments. Active planning experiments demonstrate the superior performance of our framework compared to implicit reconstruction baselines that do not consider semantics and a semantic-targeted approach using an explicit map representation. We also show that considering exploration is crucial for semantic-targeted view planning in challenging scenes to enable finding occluded objects of interest. One limitation of our current work is the assumption of access to accurate semantic labels. In the presence of noisy semantics, future work will consider integrating the uncertainty of semantic rendering in our planning pipeline.

\bibliographystyle{IEEEtranSN}
\footnotesize
\bibliography{reference}

\end{document}